\documentclass[11pt, a4paper, logo, copyright]{googledeepmind}

\usepackage[authoryear, sort&compress, round]{natbib}
\usepackage[inkscapeformat=png]{svg}

\usepackage[most, breakable, skins]{tcolorbox}

\tcbuselibrary{skins}
\usepackage{lipsum}
\usepackage{siunitx}
\usepackage{tabularx}
\usepackage{afterpage}
\usepackage{booktabs}
\usepackage{subcaption}

\usepackage{threeparttable}
\usepackage{makecell}
\usepackage{multirow}
\usepackage{multicol}
\usepackage{array}
\usepackage{epigraph}
\usepackage{float}
\usepackage{listings, listings-rust}
\usepackage{fontawesome5}
\usepackage{hyperref}
\usepackage{amssymb,graphicx}
\usepackage[dvipsnames]{xcolor}
\usepackage{hyperref}
\usepackage[capitalise, noabbrev]{cleveref}
\usepackage{longtable}
\usepackage{graphicx}
\usepackage{ragged2e}
\usepackage{pdflscape}
\usepackage{adjustbox}
\usepackage[section]{placeins}  
\usepackage{tikz}
\usetikzlibrary{shapes.geometric, arrows, positioning, fit}
\usepackage{tcolorbox} 
\usepackage{listings}
\usepackage{xspace}
\usepackage{caption} 
\usepackage{pifont}

\usepackage{titlesec}
\titlespacing*{\section}{0pt}{*1}{*0.5}
\titlespacing*{\subsection}{0pt}{*0.8}{*0.4}

\newcommand{\ra}[1]{\renewcommand{\arraystretch}{#1}}

\newcommand*{\colorboxed}{}
\def\colorboxed#1#{%
  \colorboxedAux{#1}%
}
\newcommand*{\colorboxedAux}[3]{%
  \begingroup
    \colorlet{cb@saved}{.}%
    \color#1{#2}%
    \boxed{%
      \color{cb@saved}%
      #3%
    }%
  \endgroup
}

\newcommand{\evalname}{\textsl{SimpleQA Verified}\xspace}

\title{\evalname: A Reliable Factuality Benchmark to Measure Parametric Knowledge}

\correspondingauthor{lukhaas@google.com}

\newcommand{\bGdm}{$\mathbin{\Diamond}$}

\newcommand{\bResearch}{\ding{171}}

\author[\bGdm]{Lukas Haas}
\author[\bResearch]{Gal Yona}
\author[\bGdm]{Giovanni D'Antonio}
\author[\bResearch]{Sasha Goldshtein}
\author[\bGdm]{Dipanjan Das}

\affil[\bGdm]{Google DeepMind}
\affil[\bResearch]{Google Research}

\begin{abstract}

We introduce \evalname, a 1,000-prompt benchmark for evaluating Large Language Model (LLM) short-form factuality based on OpenAI's SimpleQA. It addresses critical limitations in OpenAI's benchmark, including noisy and incorrect labels, topical biases, and question redundancy. \evalname was created through a rigorous multi-stage filtering process involving de-duplication, topic balancing, and source reconciliation to produce a more reliable and challenging evaluation set, alongside improvements in the autorater prompt. On this new benchmark, Gemini 2.5 Pro achieves a state-of-the-art F1-score of 55.6, outperforming other frontier models, including GPT-5. This work provides the research community with a higher-fidelity tool to track genuine progress in parametric model factuality and to mitigate hallucinations. The benchmark dataset, evaluation code, and leaderboard are available at: \url{https://www.kaggle.com/benchmarks/deepmind/simpleqa-verified}.

\end{abstract}

\begin{document}

\maketitle

\section{Introduction}
\label{sec:intro_revised}

The capacity of Large Language Models (LLMs) to generate factually accurate information is vital for their utility and a primary concern for widespread adoption. Inaccurate or "hallucinated" outputs erode user trust and still present barriers, particularly in critical enterprise applications where factual reliability is non-negotiable. Consequently, the rigorous evaluation of model factuality has become a central focus of AI research, driving the need for benchmarks that can accurately measure and differentiate the capabilities of state-of-the-art systems.

To meet this need, researchers have developed diverse evaluation paradigms. Many prominent approaches assess a model's ability to reason over externally provided information; this includes grounding evaluations, which test factuality with respect to a given context~\citep{jacovi2025factsgrounding, rashkin2022measuringattribution}, and retrieval-augmented benchmarks which evaluate the use of search tools to access fresh or real-time information beyond the model's training data~\citep{vu2023freshllms, kasai2024realtimeqa, krishna2025factfetch, yang2024crag}. Another line of work targets the complexity of long-form generation, where factual accuracy must be maintained across extended responses~\citep{chen2023felm, pan2023factchecking, wei2024longformfactuality, jacovi2025factsgrounding, song2024veriscore}. Distinct from these, our work concentrates on a model's ability to recall facts directly from its internal parameters. This domain of \textit{parametric factuality}, typically measured with short-form question-answer (QA) formats, isolates the model's stored knowledge from external aids and can easily and reliably be verified with LLM autoraters~\citep{wei2024measuringshortformfactualitylarge, min-etal-2023-factscore, pan2023factchecking}.

The landscape of benchmarks for this specific form of QA factuality has evolved from early standards like TriviaQA~\citep{joshi-etal-2017-triviaqa}, Natural Questions~\citep{kwiatkowski-etal-2019-natural}, and TruthfulQA~\citep{lin2022truthfulqa}. While instrumental in their time, these datasets have become saturated by the performance of modern LLMs, limiting their ability to provide a meaningful signal on frontier models. In response to this, OpenAI released \textit{SimpleQA} in late 2024, a more challenging benchmark for short-form, parametric factuality that quickly became an industry standard~\citep{wei2024measuringshortformfactualitylarge}.

Despite its difficulty, the utility of SimpleQA is compromised by significant limitations; benchmark questions are derived from a narrow distribution of source documents due to substantial human rater biases. Additionally, SimpleQA suffers from incorrect ground truths and disproportionally leans toward specific topics and question formats. The dataset contains a high degree of redundancy with semantically similar or lexically overlapping questions. These issues create a noisy evaluation signal, making it difficult to discern whether performance gains stem from genuine improvements in factual recall or from models overfitting to the benchmark's specific quirks.

To address these shortcomings, we introduce \evalname, a cleaner, more reliable and robust benchmark of 1,000 prompts filtered and modified from the original SimpleQA dataset (see \Cref{tab:examples}). \evalname was curated through a multi-stage process involving removing duplicate sources, semantic and TF-IDF de-duplication, re-balancing of topic and answer-type distributions to ensure question diversity, reconciliation of conflicting sources to verify ground truths, and a filtering step to align reference URLs in our benchmark with the crawling preferences of their web publishers.

By applying this rigorous methodology and further improving the autorater prompt, we provide a more reliable benchmark for measuring parametric factuality, designed to be evaluated without any tools (i.e. search). Our results on a suite of leading models show that Gemini 2.5 Pro~\citep{comanici2025gemini25pushingfrontier} leads performance on both the original SimpleQA as well as \evalname benchmarks, outperforming even more recently released frontier models including GPT-5~\citep{openai2025gpt5card} and Claude Opus 4~\citep{anthropic2025claude4card}. We release \evalname to the community to enable more precise and trustworthy assessments of LLM factuality, fostering progress toward more reliable AI systems.

\begin{table*}[ht]\centering
\caption{Examples from \evalname.}
\label{tab:examples}
\scriptsize
\ra{1.3}
\resizebox{0.99\linewidth}{!}{
\begin{tabular}{
p{5.5cm}
p{3.5cm}
l
l
c
c
}
\toprule
\textbf{Problem} & \textbf{Answer} & \textbf{Topic} & \textbf{Answer Type} & \textbf{Reasoning} & \textbf{Multi-Step}\\ \midrule
 On what day, month, and year was Algerian artist Mohammed Racim born? & June 24, 1896 & Art & Date & NO & NO\\ \\
 In how many games did Matija Radović appear for the Hofstra Pride during the 2017-18 season? &  25 & Sports &  Number & NO & NO\\ \\
 What is the longitude of Lilongwe in decimal format? &  33.7738 (acceptable range: anything between 33.7586 and 33.8022) & Geography &  Number & NO & NO \\ \\
 From which university did David Hibbett receive his Bachelor of Arts degree? &  University of Massachusetts Amherst & Other & Place & NO & NO\\ \\
 Which famous drummer opened a Wahoo's Fish Taco restaurant in Norco, California, in 2004? & Travis Barker &  Music & Person & YES & NO\\ \\
 What was the age gap between George Frederic Watts and his first wife, Ellen Terry? & 30 years. & Art & Number & NO & YES
\\ \bottomrule
\end{tabular}
}
\end{table*}

\section{\evalname}
\label{sec:data}

The \evalname benchmark dataset was carefully curated to correct a range of issues present in the original SimpleQA benchmark~\citep{wei2024measuringshortformfactualitylarge}, likely caused by misaligned incentives of human raters who created the original benchmark prompts and and missing subsequent filtering and validation stages. The complete methodology how \evalname was created is outlined in the following subsections in the order they were performed. The authors carried out these tasks themselves. Table \ref{tab:processing} illustrates our methodology and  shows how many questions remain after each processing step.

\begin{table*}[ht]\centering
\caption{Processing steps to create \evalname.}
\label{tab:processing}
\scriptsize
\ra{1.3}
\resizebox{0.99\linewidth}{!}{
\begin{tabular}{
p{7.5cm}
c
c
c
}
\toprule
\textbf{Processing Stage} & \textbf{Dataset Size} & \textbf{Cum. Samples Removed} & \textbf{Gemini 2.5 Pro F1-Score}\\ \midrule
 SimpleQA~\citep{wei2024measuringshortformfactualitylarge} & 4,326 & 0.0\% & $55.1\%$\\
 \midrule
 Ensuring Unique Source Documents & 3,095 & $-28.5\%$ & $54.3\%$\\
 Removing Highly Similar Questions (Embeddings) & 2,871 & $-33.6\%$ & $54.4\%$\\
 Removing Highly Similar Questions (TF-IDF) & 2,664 & $-38.4\%$ & $54.4\%$\\
 Respecting Web Publisher Choices & 1,855 & $-57.1\%$ & $55.0\%$\\
 Ensuring Diversity Across Answer Types and Topics & 1,218 & $-71.8\%$ & $54.0\%$\\
 Reconciliation of Conflicting Sources (Non-Numeric) & 1,117 & $-74.2\%$ & $56.1\%$\\
 Reconciliation of Conflicting Sources (Numeric) & 1,073 & $-75.2\%$ & $56.5\%$\\
 Rewriting Numeric Ground Truths with Acceptable Ranges & 1,073 & $-75.2\%$ & $58.4\%$\\
 \midrule
 \evalname (after Increasing Benchmark Headroom) & 1,000 & $-76.9\%$ & $55.6\%$
\\ \bottomrule
\end{tabular}
}
\end{table*}

\subsection{Ensuring Unique Source Documents}\label{sec:2_1}
SimpleQA~\citep{wei2024measuringshortformfactualitylarge} was created as a challenging factuality benchmark in response to older short-answer factuality benchmarks like TriviaQA~\citep{joshi-etal-2017-triviaqa} and Natural Questions~\citep{kwiatkowski-etal-2019-natural} becoming saturated. As a result, SimpleQA contains predominantly questions asking for niche (or so-called ``tail'') knowledge. To ensure these niche questions are representative of topics that users might ask about, and are not over-sampled with respect to the preferences of a single human rater, we filtered our dataset to ensure that no two questions would share the same reference URL. Every question in the original SimpleQA dataset contains at least two reference URLs, each provided by a distinct human rater. This reduced the dataset size \textbf{from 4,326 to 3,095 questions} ($-28.5\%$).

To provide maximal headroom for AI models to hill-climb on \evalname, for a given set of questions which shared the same reference URL, we randomly sampled a question which all of GPT-4o~\citep{openai2024gpt4ocard}, Gemini 2.0 Flash~\citep{gemini_2_0}, and Claude 3.7 Sonnet~\citep{claude_3_7} answered incorrectly. If no question was answered incorrectly by all three models, we chose a question to keep in our dataset at random.

\subsection{Removing Highly Similar Questions}

\begin{table*}[ht]\centering
\caption{Examples of highly similar questions from the original SimpleQA benchmark.}
\label{tab:deduplication}
\scriptsize
\ra{1.3}
\resizebox{0.99\linewidth}{!}{
\begin{tabular}{
c
p{1.5cm}
p{6.5cm}
p{4.0cm}
c
}
\toprule
\textbf{\# of Cases} & \textbf{Method} & \textbf{Prompt} & \textbf{Answer}  & \textbf{Kept} \\ \midrule
 \multirow{3}{2em}{\vspace{-3em}119} & \multirow{3}{3em}{\vspace{-3em}Embeddings} & What day, month, and year was the municipality of Tipacoque, Boyacá, Colombia, created? & November 28th, 1968 &  YES \\
 & & In which year was the municipality of Motavita, Boyacá, Colombia, founded? & 1816 &  NO \\
 & & What year was the municipality of Turbo, Antioquia, Colombia, founded? & 1840 &  NO \\
 \midrule
 \multirow{3}{2em}{\vspace{-3em}8} & \multirow{3}{3em}{\vspace{-3em}TF-IDF} & What is the name of the district with the Regional Transport Office (RTO) code SK-06 in Sikkim, India? & Soreng &  YES \\
 & & What is the Regional Transport Office (RTO) code for the Androth location in Lakshadweep, India? & LD-04 &  NO \\
 & & What is the Regional Transport Office (RTO) code for the Phek district location in Nagaland, India? & NL-08 &  NO \\
 \midrule
 \multirow{3}{2em}{\vspace{-3em}7} & \multirow{3}{3em}{\vspace{-3em}TF-IDF} & In which episode and season of South Park does Aunt Jemima first appear? Give me the number and title. & Episode 2: Gluten Free Ebola, Season eighteen &  YES \\
 & & In which season and episode of South Park does Randy become a chef at South Park Elementary? & Season 14, "Crème Fraîche" &  NO \\
 & & In which season and episode of South Park does Stan ask why dogs have cold noses? & Season 2 Episode 3: "Ike's Wee Wee" &  NO
\\ \bottomrule
\end{tabular}
}
\end{table*}

SimpleQA~\citep{wei2024measuringshortformfactualitylarge} contains many highly similar questions which often seem to stem from the same human rater (no rater IDs were provided in OpenAI's release), likely a result of misaligned rating incentives. Table \ref{tab:deduplication} illustrates such cases; for example, SimpleQA contains 119 questions (2.7\% of dataset) asking about the founding dates of different Colombian municipalities. We use a combination of semantic and TF-IDF de-duplication to ensure questions in \evalname are meaningfully distinct from each other and challenge AI models across diverse domains.

\paragraph{Semantic De-Duplication with Gemini Embeddings.} In a first step, we compute Gemini Embeddings~\citep{lee2025geminiembeddinggeneralizableembeddings} for all prompts in SimpleQA. We find that a cosine similarity cutoff of 0.77 works well to identify questions which are unreasonably similar and can be de-duplicated. We use the same difficulty filter as in Section \ref{sec:2_1} to ensure we do not retain questions which can be answered by all leading frontier language models. De-duplicating the dataset semantically using embeddings reduced the number of samples in our dataset \textbf{from 3,095 to 2,871} ($-7.2\%$).

\paragraph{De-Duplication Using TF-IDF.} In a subsequent step, we compute a TF-IDF matrix~\citep{tf_idf} to find unreasonably similar questions using exact word matches. Again, we compute a cosine similarity between vectors, and manually review all cases which have similarities above a threshold of 0.4, retaining questions which are difficult to answer by leading frontier models. This further reduces our dataset size \textbf{from 2,871 to 2,664} ($-7.2\%$).

\subsection{Respecting Web Publisher Choices}
In some instances, web publishers may choose to manage site access using instructions in their \lstinline{robots.txt} file. For example, \href{https://developers.google.com/search/docs/crawling-indexing/google-common-crawlers}{Google-Extended} is a standalone product token that web publishers can use to manage whether content Google crawls from their sites may be used for training future generations of Gemini models that power Gemini Apps and Vertex AI API for Gemini and for grounding in Gemini Apps and Grounding with Google Search on Vertex AI. Other model providers such as OpenAI and Anthropic also offer controls to web publishers available through instructions in \lstinline{robots.txt} files.

A number of the reference URLs provided in SimpleQA~\citep{wei2024measuringshortformfactualitylarge} were associated with web publishers that have adopted these controls. SimpleQA contains niche questions where information addressing them might only be found on very few websites, and, as a result, a decision was made to remove questions from the dataset whose reference URLs were associated with web publishers that have adopted the types of controls set out above from Google, OpenAI or Anthropic. Removing these questions reduces the dataset \textbf{from 2,664 to 1,855} ($-30.4\%$) examples.

\subsection{Ensuring Diversity Across Answer Types and Topics}
\begin{figure}[ht]
    \centering
    \begin{minipage}{0.48 \textwidth}
        \centering
        \includegraphics[width=\linewidth]{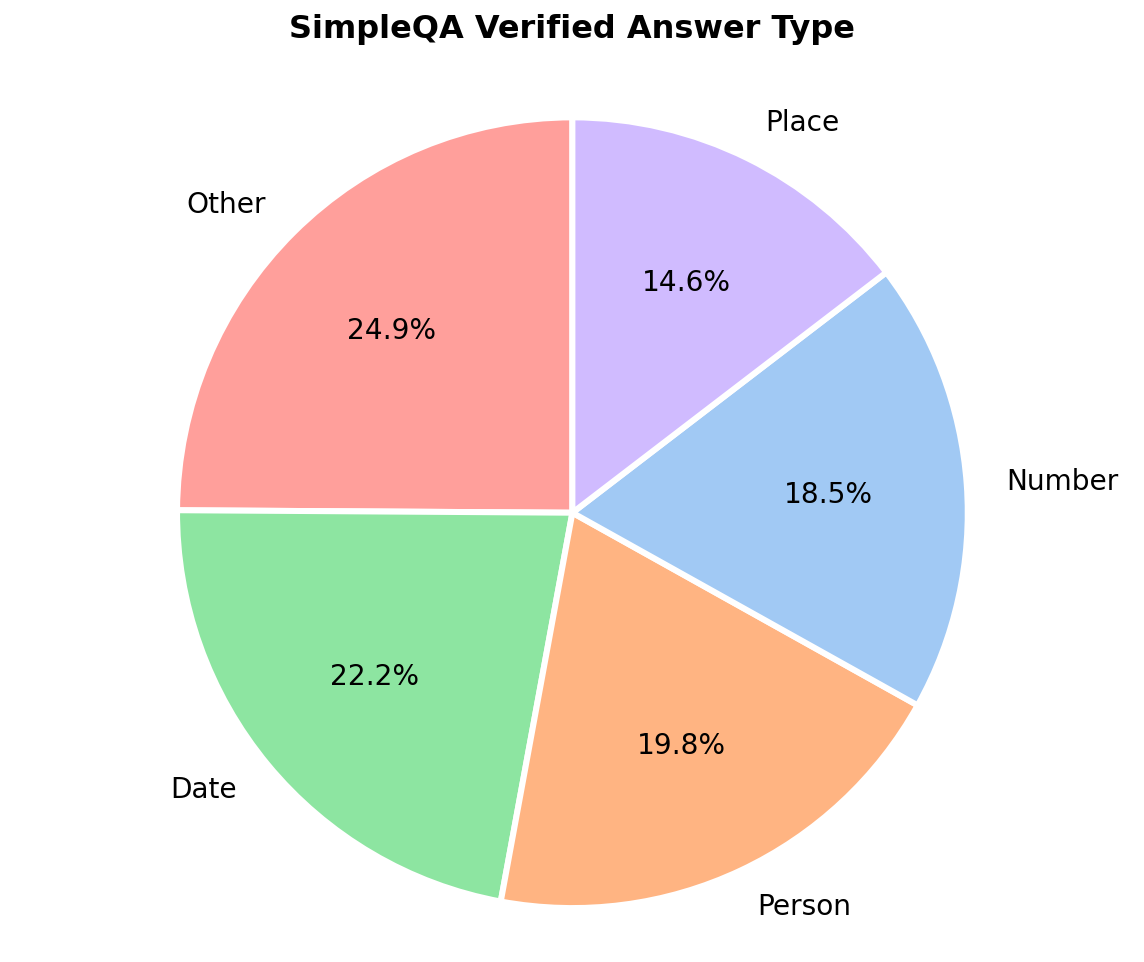}
    \end{minipage}
    \begin{minipage}{0.48\textwidth}
        \centering
        \includegraphics[width=\linewidth]{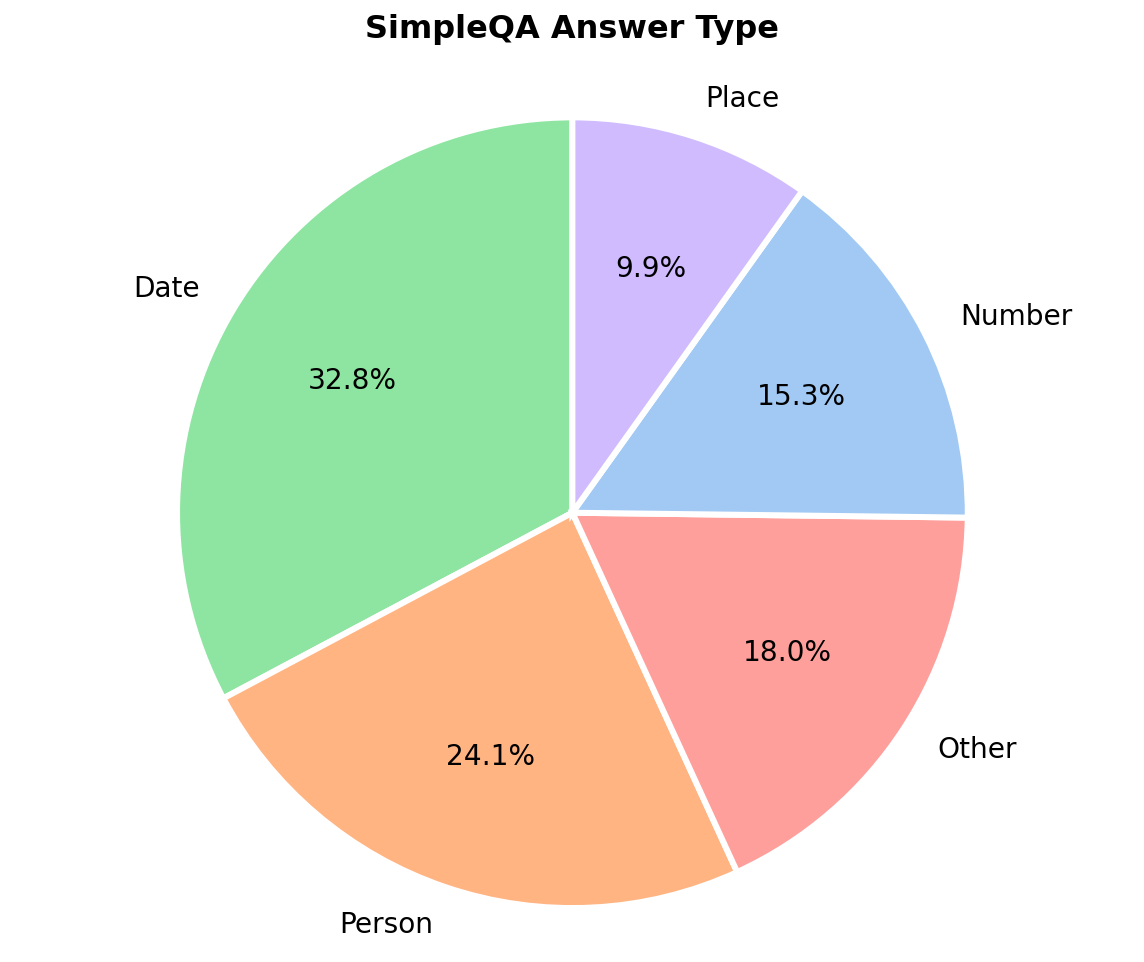}
    \end{minipage}
    \caption{Distributions of answer types as a percent of the total number of data points in \evalname and \textit{SimpleQA}. The answer type classification was initially performed by~\citet{wei2024measuringshortformfactualitylarge}.}
    \label{fig:distr_answer}
\end{figure}

\begin{figure}[ht]
    \centering
    \begin{minipage}{0.52 \textwidth}
        \centering
        \includegraphics[width=\linewidth]{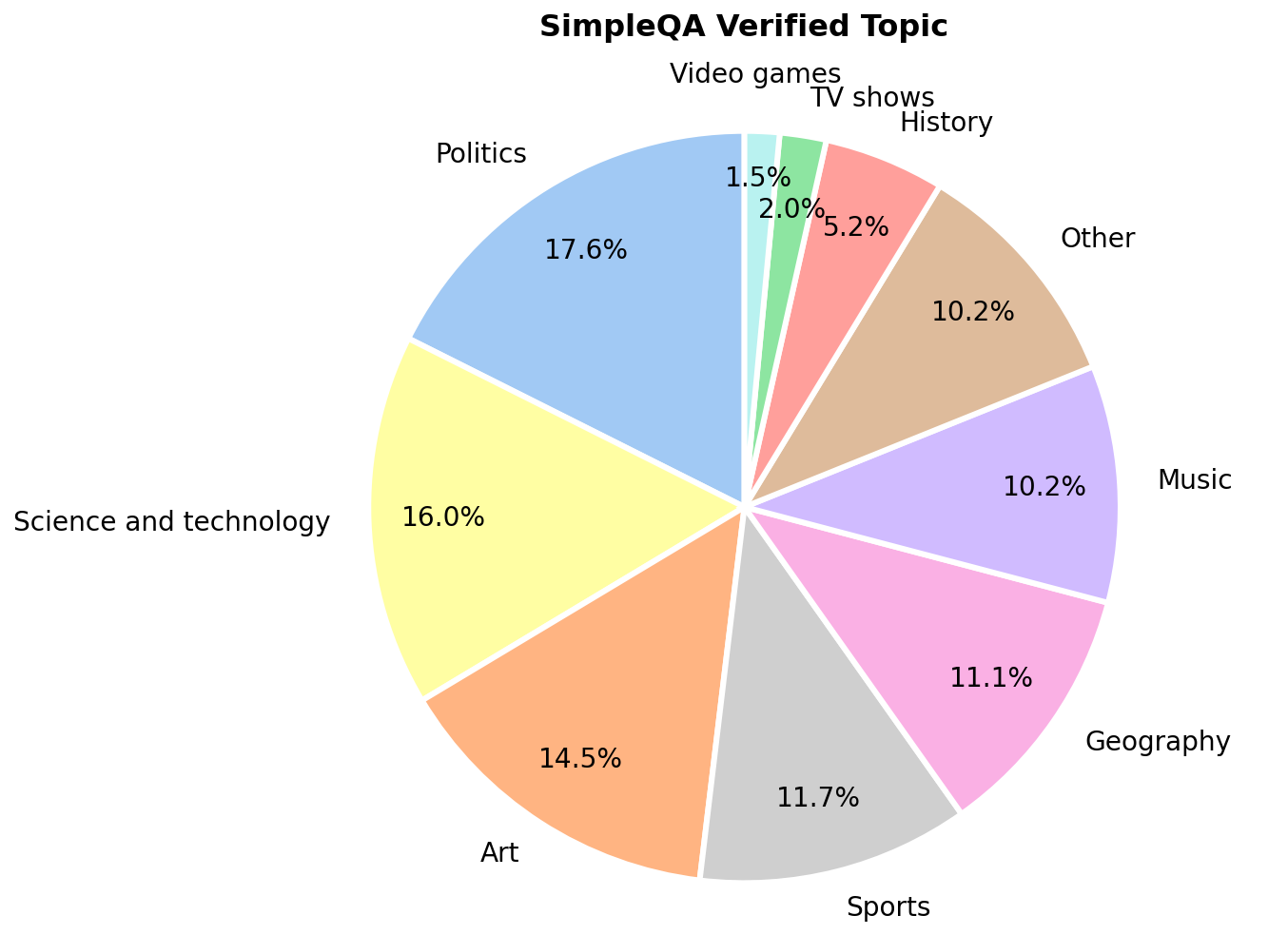}
    \end{minipage}
    \begin{minipage}{0.46\textwidth}
        \centering
        \includegraphics[width=\linewidth]{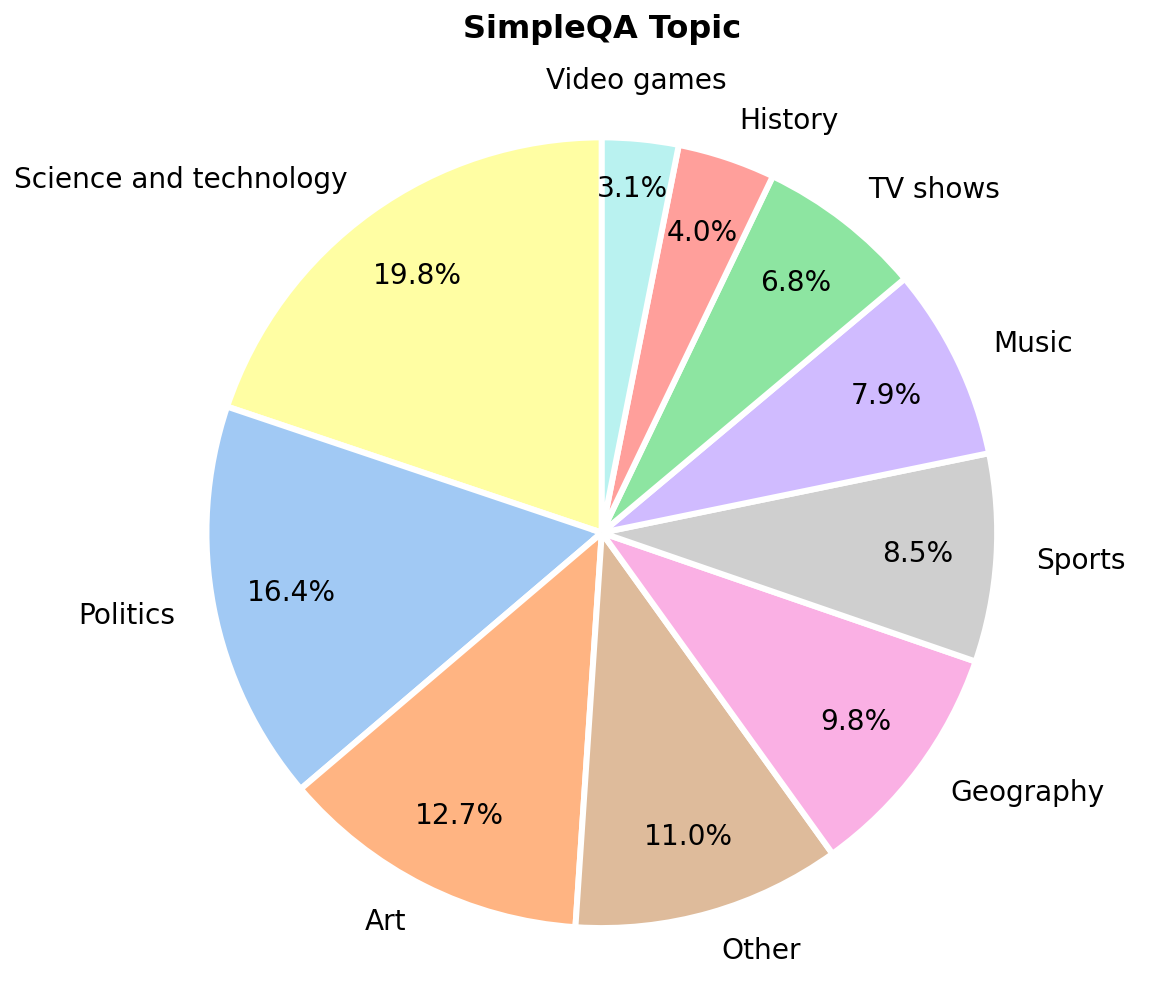}
    \end{minipage}
    \caption{Distributions of question topics as a percent of the total number of data points in \evalname and \textit{SimpleQA}. The topic classification was initially performed by~\citet{wei2024measuringshortformfactualitylarge}.}
    \label{fig:distr_topic}
\end{figure}

In the original SimpleQA~\citep{wei2024measuringshortformfactualitylarge} paper, certain topics and answer types are overrepresented, and the distribution of answer types can skew benchmark results. For instance, 32.8\% of questions in SimpleQA require a date and 24.1\% require a person's name as the answer (see Figure \ref{fig:distr_answer}). A robust factuality benchmark of parametric knowledge must therefore account for such distributions to avoid unfairly penalizing models with specific weaknesses, such as in date processing. Similarly, when it comes to  question topics, SimpleQA over-indexes on science and technology domains (see Figure~\ref{fig:distr_topic}). \pagebreak

\vspace{2em}

For \evalname, we sub-sample our remaining samples with the primary goal balancing answer types and the secondary goal of balancing question topics. When choosing which questions to remove within a given answer type or topic category, we again adversarially determine the set of hardest questions using the method described in Section~\ref{sec:2_1}. Using our filtering technique, we obtain a set of 1,218 questions which is still sizeable enough to allow for meaningful statistics, while also ensuring \evalname is a well-balanced factuality benchmark with diverse questions. As part of the described process, our dataset shrinks \textbf{from 1,855 to 1,218} ($-34.3\%$) examples.\\

\vspace{2em}

\pagebreak

\subsection{Reconciliation of Conflicting Sources}
Using an ensemble of search-augmented models and the reference URLs provided by the original SimpleQA~\citep{wei2024measuringshortformfactualitylarge} benchmark, we review potentially conflicting sources in our dataset. For any conflicting sources we identify, we treat numeric (\lstinline{answer_type=Number}) and non-numeric answer types differently.

\paragraph{Non-Numeric Answer Types.}
We remove questions from the dataset which are clearly ambiguous or where sources point to more than one distinct answer. Additionally, where possible, we correct both answers and reference URLs. This reduces our dataset further \textbf{from 1,218 to 1,117} ($-8.3\%$) samples.

\paragraph{Numeric Answer Types.}
For questions asking for a numeric answer, we employ a prompted autorater with code execution to classify whether all sources (search augmented LLM and reference URLs) point to answers  within a 5\% margin of error with respect to the ground truth of SimpleQA. We remove all questions which have answers outside of the 5\% margin of error across all sources. This lowers the dataset size \textbf{from 1,117 to 1,073} ($-3.9\%$).

\subsection{Increasing Benchmark Headroom}
\label{sec:headroom}
By cleaning the benchmark dataset and removing incorrectly labeled, ambiguous, or questions with irreconcilable sources, the benchmark becomes easier to solve. To ensure model developers have a similar headroom to hill-climb on for \evalname, we filter the 1,000 most difficult questions from our remaining samples. We do this by again discarding questions which all leading frontier models solve correctly (see Section \ref{sec:2_1}). This brings down the dataset size \textbf{from 1,073 to 1,000} ($-6.8\%$).

\subsection{Manual Review \& Metadata Enrichment}
Having selected a final set of 1,000 questions for \evalname, we perform a range of manual checks and modifications:
\begin{enumerate}
    \item \textbf{URL Cleaning:} a range of reference URLs in the original benchmark are either invalid or irrelevant (i.e. a link to an NFT). We remove these URLs or correct them where possible.\\
    \item \textbf{Source quality:} to evaluate source diversity, we examined the most frequent domains in the reference URLs. The results show a strong prevalence of established encyclopedias, indicating that \evalname relies on credible and varied sources rather than a narrow set of overrepresented domains.\\
    \item \textbf{Date precision:} some questions with a \lstinline{Date} answer type ask for just the month and year, or a date, month and year, but the provided ground truth is either too specific or not specific enough. By checking the sources, we correct these labeling mistakes.\\
    \item \textbf{Metadata enrichment:} using a prompted classifier, we detect which questions might require reasoning capabilities (i.e. calculating the difference between two dates) and which questions are multi-step (requiring multiple sequential information steps) in nature~\citep{yang2018hotpotqa, geva2021didaristotleuselaptop}. $3.7\%$ of questions are classified as requiring reasoning and $7.3\%$ are multi-step. This metadata is added to the \evalname dataset.
\end{enumerate}

\pagebreak

\section{Metrics \& Autorater Improvements}

We use the same grading scheme and metrics as proposed in \citet{wei2024measuringshortformfactualitylarge}. Specifically, each response is automatically graded as either \emph{correct}, \emph{incorrect} or \emph{not attempted}, and the final metric is obtained as the harmonic mean (F1-Score) of \emph{overall correct} (total number of questions that were answered correctly) and \emph{correct given attempted} (number of questions that were answered correctly out of those that were attempted). However, we identify several areas for improvement in the automatic grading process itself. In the rest of this section we detail this process and overview the proposed modifications to the prompted autorater.

\subsection{Methodology}
\label{sec:autorater_methodology}

Despite the focused nature of the questions in \evalname, modern AI systems still often produce longer and more elaborate responses. These typically contain more information than is necessary to strictly answer the provided question. Our objective is to ensure the prompted autorater can be robustly applied to grade such responses. To stress-test this, we perform an automatic analysis designed to surface responses for which the existing autorater, proposed in \cite{wei2024measuringshortformfactualitylarge}, currently exhibits high uncertainty about their grade. We first sample responses to \evalname from an ensemble of models. We then grade each response using the existing prompted autorater $N=10$ times at high temperature ($T=2.0$). Finally, we manually review the responses for which disagreement between the $N$ sampled grades was highest. We find several recurring patterns in models responses that frequently confuse the autorater with respect to the assigned grade. These are summarized in \autoref{tab:autorater_stability_loss_buckets}.\\

\begin{table*}[ht]\centering
\caption{Recurring patterns of responses  for which the existing prompted autorater  exhibited high disagreement, highlighting possible headroom for  improvements in the grading process. The top half of the table depicts cases where the disagreement is about the answer's \emph{correctness}; in the bottom half, the disagreement is about whether the answer should be considered attempted.
}
\label{tab:autorater_stability_loss_buckets}
\tiny
\ra{1.3}
\resizebox{0.99\linewidth}{!}{
\begin{tabular}{
p{2cm}
p{3cm}
p{1cm}
p{6cm}
}
\toprule
\textbf{Pattern} & \textbf{Question} &  \textbf{Answer} & \textbf{Predicted Answer} \\  \midrule
 Predicted answer is correct, but adds extra information or corrects an assumption in the question & Who was the governor of Assam from 16 February 1949 to 27 May 1949? & Sri Prakasa & The governor of Assam from 16 February 1949 to 27 May 1949 was Sri Prakasa. He took office on 16 February 1949 and served until 27 May 1950. The period you asked about falls within his tenure. \\ \\
 Predicted answer provides an approximately correct answer to a numeric question &  By 1913, how many stars could astronomer Annie Jump Cannon classify per hour? & 200 & By 1913, Annie Jump Cannon could classify three stars per minute, which translates to 180 stars per hour. \\  \\
 \midrule
 \\
Predicted answer claims the information doesn't exist or is hard to find  &  What is the full name of the Colombian soccer player Juan Guzmán born in 1988? & Juan Pablo Guzmán Perdomo & I was unable to find a Colombian soccer player named "Juan Guzmán" born in 1988 in the major soccer databases I have access to.
It's possible that:
* His name is slightly different.
* He played at a lower or amateur level that isn't widely documented online.
* He is not primarily known as a professional soccer player.
If you have any other details about him (like the teams he played for, his position, etc.), that might help to narrow down the search.  \\ \\
Predicted answer is a soft refusal, hedge or pushback
 & How many years did Burt work in the Optics and Design department in the show "Severance"? & 7 & The show "Severance" doesn't explicitly state the exact number of years Burt worked in the Optics and Design department.
However, it's made clear that he has been there for a very long time, serving as the department chief. His retirement in the series signifies a long tenure at Lumon. So, while we know it was a significant period, a precise number of years isn't provided.
\\  \\ \bottomrule
\end{tabular}
}
\end{table*}
\pagebreak

\subsection{Autorater Improvements}
\label{sec:autorater_improvements}

With these findings in mind, we make the following modifications to the automatic grading process:

\begin{itemize}
    \item \textbf{Improved grading of answers to numeric questions}: To ensure the robustness of grading answers to numeric questions, we remove the generic instruction in the prompted autorater (\emph{``the predicted answer needs to be correct to the last significant figure in the gold answer''}) and instead explicitly specify the range of answers that should be considered correct as part of the ground truth answer. For example, given the question ``How many UN Volunteers served in 54 United Nations missions, agencies, funds, and programs across the globe in 2019?'' the ground truth answer 8,282 will be re-written as ``\emph{8282 (acceptable range: anything between 8199 and
    8365)}''; see additional examples in \autoref{tab:numeric_rewrite_examples}. The new prompted autorater is instructed to consider any predicted answer within this range as correct. The range is chosen on an input-specific basis; see the taxonomy detailed in \autoref{tab:numeric_rewrite_taxonomny}.
\end{itemize}

\begin{table*}[th]\centering
\caption{Rewriting the ground truth answer to explicitly specify the allowed margin-of-error using the taxonomy in \autoref{tab:numeric_rewrite_taxonomny} leads to a more robust grading of approximate numeric answers.}
\label{tab:numeric_rewrite_examples}
\scriptsize
\ra{1.3}
\resizebox{0.99\linewidth}{!}{
\begin{tabular}{
p{5.5cm}
p{3.5cm}
p{5.5cm}
l
c
}
\toprule
\textbf{Problem} & \textbf{Answer} & \textbf{Predicted answer} & \textbf{Original AR Grade} & \textbf{New AR Grade} \\ \midrule
 How many UN Volunteers served in 54 United Nations missions, agencies, funds, and programs across the globe in 2019? & 8,282 (acceptable range: anything between 8199 and 8365) & In 2019, a total of 8,333 UN Volunteers served in 54 United Nations missions, agencies, funds, and programs across the globe. & INCORRECT & CORRECT \\ \\
 What is the exact number of visitors who visited Edinburgh Castle in 2018? &  2,111,578 (acceptable range: anything between 2,006,000 and 2,217,157) & According to the official figures released by the Association of Leading Visitor Attractions (ALVA), the exact number of visitors to Edinburgh Castle in 2018 was 2,111,532. &  INCORRECT & CORRECT \\ \\
 What was the length of Prafullachandra Natwarlal Bhagwati's tenure as the Chief Justice of India, in years and days? &  1 year, 161 days (acceptable range: 1 year, anything between 160 and 162 days) & Prafullachandra Natwarlal Bhagwati's tenure as the 17th Chief Justice of India was from July 12, 1985, to December 20, 1986. The total length of his tenure was 1 year and 162 days. &  INCORRECT & CORRECT  \\ \\
In 1989, how many kilometers long was the Cape Town Cycle Tour, formerly known as the Cape Argus Cycle Tour?
 & 105 km (acceptable range: anything between 103 and 107 km) & In 1989, the Cape Argus Cycle Tour was shortened to 103 kilometers due to extreme heat.  & INCORRECT & CORRECT
\\ \bottomrule
\end{tabular}
}
\end{table*}

\vspace{2em}

\begin{table*}[th]\centering
\caption{For grading numeric answers, we specify the allowed margin-of-error on a question-specific basis using the following taxonomy.}
\label{tab:numeric_rewrite_taxonomny}
\tiny
\ra{1.3}
\resizebox{0.99\linewidth}{!}{
\begin{tabular}{
p{2.0cm}
p{7.5cm}
l
}
\toprule
\textbf{Category} & \textbf{Logic} & \textbf{\# observations (\%)}  \\ \midrule
 Exclude\newline(Not Numeric / Error) & Ground truth answer is not a meaningful quantity, but rather an identifier, code, or a clear data error/nonsensical value. & 16 (8\%)  \\ \\
 Exact Match Required &  For small discrete counts (integers $\leq 50$) of enumerable items, or quantities that are inherently exact and not subject to approximation (e.g., specific identifiers, fixed values in a sequence or definition, fundamental mathematical constants) & 89 (44\%)  \\ \\
 Small Margin of Error\newline(approx. $\pm1\%$) &  For medium-sized quantities (generally 51 to 10,000) that represent continuous measurements or common counts where slight variation is acceptable (e.g., lengths, areas, medium populations, votes, coordinates). & 84 (42\%)  \\ \\
Large Margin of Error\newline(approx. $\pm5\%$)
 & For very large quantities (generally > 10,000) that represent large-scale measurements or aggregate data where larger estimation margins are common (e.g., very large populations, national budgets, extreme vote counts). & 12 (6\%)
\\ \bottomrule
\end{tabular}
}
\end{table*}

\begin{itemize}
    \item \textbf{Clarified guidelines around direct answers and hedging}: We clarify the guidelines in the prompt around two specific areas. First, we emphasize that for the purpose of the assigned grade, only the part of the predicted answer that directly answers the question should be taken into account, meaning any additional information should be ignored. Second, we clarify the guidelines around hedged responses that contain multiple candidates for the final answer. The new guidelines require that such a response eventually commits to one of the candidates as the more likely answer -- and that answer should be judged for correctness. Otherwise, the predicted answer should be considered as not attempted. This is important to ensure that the strategy of iterating various possible guesses (one of which may be correct) is consistently judged as \emph{not attempted} and thus is not beneficial.\\

    \item \textbf{Diversified examples of punting styles}: In the few-shot examples included in the prompt we incorporate demonstrations of additional punting styles, to ensure that such predicted answers are consistently graded as \emph{not attempted}.
\end{itemize}

The full prompt is provided in \autoref{app:measurement}; differences from the original prompt in \cite{wei2024measuringshortformfactualitylarge} are highlighted in red. The total number of words increases by 15\% (963 → 1,124).\\

\section{Results}
\label{sec:results}

\Cref{tab:results} contains the results of a set of commercially available frontier models on \evalname. The tested models include \textit{Gemini 2.5 Flash Lite}, \textit{Flash}, and \textit{Pro} from the Gemini model family~\citep{comanici2025gemini25pushingfrontier}, \textit{GPT 4o}, \textit{4.1}, \textit{o3}, \textit{o4} as well as \textit{GPT 5}, \textit{5 Mini}, and \textit{5 Nano} from the GPT family~\citep{openai2024gpt4ocard,openai2025gpt4-1card,openai_learning_to_reason,openai2025gpt5card}, \textit{Sonnet 4} and \textit{Opus 4} from Anthropic's Claude models~\citep{anthropic2025claude4card}, and the latest \textit{DeepSeek R1}~\citep{deepseekr1} version. All models are evaluated with their standard API parameters and without tools. Enabling tools on \evalname results in near perfect performance, emphasizing that \evalname should be employed for measuring parametric factuality only.

In our results, we report the same metrics computed in \cite{wei2024measuringshortformfactualitylarge} and use \lstinline{gpt-4.1-2025-04-14} as an autorater. Additionally, we measure the difference in scores between \evalname and \textit{SimpleQA} on all models. The results of \textit{GPT 4o}, \textit{Claude Opus 4}, \textit{Claude Sonnet 4} are statistically significantly worse on \evalname compared to \textit{SimpleQA}, whereas \textit{o4-mini}'s score improves. On average, model performance on \evalname is almost exactly the same as on \evalname – our cleaning process described in \Cref{sec:data} removes erroneous and ambiguous questions which makes the benchmark easier. This is balanced by adversarially selecting a subset of challenging samples in \Cref{sec:headroom}. Gemini 2.5 Pro leads \textit{SimpleQA} and \evalname in both Accuracy and Accuracy Given Attempted (Acc.|Attempted) metrics, resulting in the highest F1-Score among frontier models.

\begin{table}[t]\centering
\caption{\evalname results across the key metrics also computed in  \cite{wei2024measuringshortformfactualitylarge}. All results are reported in percent (\%) unless stated otherwise.}\label{tab:results}
\scriptsize
\ra{1.2}
\resizebox{0.99\linewidth}{!}{
\begin{threeparttable}
\begin{tabular}{cl|cS[table-format=-1.1, table-align-text-post=false]|cccS[table-format=2.1, table-align-text-post=false]}\toprule
\textbf{Rank} & \textbf{Model} & \textbf{F1-Score} & $\Delta$ \textbf{SimpleQA (\%pt)} & \textbf{Accuracy} & \textbf{Acc.|Attempted} & \textbf{Attempted} & \textbf{Hedged}\\\midrule
\textbf{1} & \textbf{Gemini 2.5 Pro} & \textbf{55.6} & 0.5 & 55.3 & 55.9 & 98.9 & 1.1\\
2 & GPT 5 & 52.3 & 1.8 & 50.9 & 53.8 & 94.6 & 5.4\\
3 & o3 & 51.9 & 1.9 & 51.6 & 52.0 & 99.3 & 0.7\\
4 & GPT 4.1 & 39.9 & -1.0 & 39.8 & 40.1 & 99.3 & 0.7\\
5 & GPT 4o & 34.9 & -3.5{*} & 34.4 & 35.5 & 97.0 & 3.0\\
6 & DeepSeek R1 (0528) & 33.3 & 1.4 & 32.7 & 33.9 & 96.4 & 3.6\\
7 & Claude Opus 4 & 28.3 & -4.0{*} & 19.2 & 54.1 & 35.5 & 64.5\\
8 & Gemini 2.5 Flash & 28.2 & -1.4 & 27.8 & 28.7 & 96.9 & 3.1\\
9 & GPT 5 Mini & 24.6 & 1.1 & 17.3 & 42.8 & 40.4 & 59.6\\
10 & o4-mini & 23.4 & 2.9{*} & 23.0 & 23.8 & 96.5 & 3.5\\
11 & Claude Sonnet 4 & 18.7 & -4.4{*} & 12.5 & 36.9 & 33.9 & 66.1\\
12 & GPT 5 Nano & 14.4 & 0.7 & 10.2 & 24.2 & 42.2 & 57.8\\
13 & Gemini 2.5 Flash Lite & 11.1 & -0.4 & 10.2 & 12.1 & 84.0 & 16.0\\
\bottomrule
\end{tabular}
\begin{tablenotes}[flushleft]
\scriptsize
\item \textit{Notes: * $p < 0.05$; }$\Delta$\textit{ SimpleQA is the F1-Score of SimpleQA Verified minus the F1-Score of SimpleQA. Both SimpleQA Verified and SimpleQA results use the same prompted autorater (gpt-4.1-2025-04-14).}
\end{tablenotes}
\end{threeparttable}
}
\end{table}

\section{Conclusion}
\label{sec:conclusion}

This paper presents \evalname, a rigorously curated 1,000-prompt benchmark designed to address the limitations of its predecessor, SimpleQA~\citep{wei2024measuringshortformfactualitylarge}, including human rater and topical biases, incorrect labels, and question redundancy. Our comprehensive, multi-stage data curation process spanning deduplication, source reconciliation, and various filtering steps combined with enhancements to the autorater for more robust evaluation, results in a higher-fidelity tool for measuring parametric factuality. On this more challenging and reliable evaluation set, Gemini 2.5 Pro establishes a new state-of-the-art, highlighting the benchmark's ability to differentiate frontier model capabilities. By releasing the \evalname dataset, its evaluation code, and a public leaderboard, we provide the research community with a more precise instrument to track genuine progress in factual recall, discourage overfitting to benchmark artifacts, and ultimately foster the development of more trustworthy AI systems.

\section{Acknowledgments}

We would like to thank the entire Gemini Factuality team for their valuable feedback, Kaggle for hosting our leaderboard and dataset, and our sponsors Anca Dragan, Avinatan Hassidim, Fernando Pereira, Koray Kavukcuoglu, Slav Petrov, and Tulsee Doshi for their technical guidance and support.

\bibliography{main}

@misc{claude_3_7,
  author={{Anthropic}},
  title={Claude 3.7 {Sonnet} and {Claude Code}},
  year={2025},
  url={https://www.anthropic.com/news/claude-3-7-sonnet},
}

@misc{anthropic2025claude4card,
      title={Introducing Claude 4}, 
      author={{Anthropic}},
      year={2025},
      url={https://www.anthropic.com/news/claude-4}, 
}

@article{comanici2025gemini25pushingfrontier,
      title={Gemini 2.5: Pushing the Frontier with Advanced Reasoning, Multimodality, Long Context, and Next Generation Agentic Capabilities}, 
      author={{Gemini Team, Google}},
      journal={arXiv preprint arXiv:2507.06261},
      year={2025},
}

@article{chen2023felm,
      title={FELM: Benchmarking Factuality Evaluation of Large Language Models}, 
      author={Shiqi Chen and Yiran Zhao and Jinghan Zhang and I-Chun Chern and Siyang Gao and Pengfei Liu and Junxian He},
      journal={arXiv preprint arXiv:2310.00741},
      year={2023},
}

@article{deepseekr1,
      title={DeepSeek-R1: Incentivizing Reasoning Capability in LLMs via Reinforcement Learning}, 
      author={DeepSeek-AI and Daya Guo and Dejian Yang and Haowei Zhang and Junxiao Song and Ruoyu Zhang and Runxin Xu and Qihao Zhu and Shirong Ma and Peiyi Wang and Xiao Bi and Xiaokang Zhang and Xingkai Yu and Yu Wu and Z. F. Wu and Zhibin Gou and Zhihong Shao and Zhuoshu Li and Ziyi Gao and Aixin Liu and Bing Xue and Bingxuan Wang and Bochao Wu and Bei Feng and Chengda Lu and Chenggang Zhao and Chengqi Deng and Chenyu Zhang and Chong Ruan and Damai Dai and Deli Chen and Dongjie Ji and Erhang Li and Fangyun Lin and Fucong Dai and Fuli Luo and Guangbo Hao and Guanting Chen and Guowei Li and H. Zhang and Han Bao and Hanwei Xu and Haocheng Wang and Honghui Ding and Huajian Xin and Huazuo Gao and Hui Qu and Hui Li and Jianzhong Guo and Jiashi Li and Jiawei Wang and Jingchang Chen and Jingyang Yuan and Junjie Qiu and Junlong Li and J. L. Cai and Jiaqi Ni and Jian Liang and Jin Chen and Kai Dong and Kai Hu and Kaige Gao and Kang Guan and Kexin Huang and Kuai Yu and Lean Wang and Lecong Zhang and Liang Zhao and Litong Wang and Liyue Zhang and Lei Xu and Leyi Xia and Mingchuan Zhang and Minghua Zhang and Minghui Tang and Meng Li and Miaojun Wang and Mingming Li and Ning Tian and Panpan Huang and Peng Zhang and Qiancheng Wang and Qinyu Chen and Qiushi Du and Ruiqi Ge and Ruisong Zhang and Ruizhe Pan and Runji Wang and R. J. Chen and R. L. Jin and Ruyi Chen and Shanghao Lu and Shangyan Zhou and Shanhuang Chen and Shengfeng Ye and Shiyu Wang and Shuiping Yu and Shunfeng Zhou and Shuting Pan and S. S. Li and Shuang Zhou and Shaoqing Wu and Shengfeng Ye and Tao Yun and Tian Pei and Tianyu Sun and T. Wang and Wangding Zeng and Wanjia Zhao and Wen Liu and Wenfeng Liang and Wenjun Gao and Wenqin Yu and Wentao Zhang and W. L. Xiao and Wei An and Xiaodong Liu and Xiaohan Wang and Xiaokang Chen and Xiaotao Nie and Xin Cheng and Xin Liu and Xin Xie and Xingchao Liu and Xinyu Yang and Xinyuan Li and Xuecheng Su and Xuheng Lin and X. Q. Li and Xiangyue Jin and Xiaojin Shen and Xiaosha Chen and Xiaowen Sun and Xiaoxiang Wang and Xinnan Song and Xinyi Zhou and Xianzu Wang and Xinxia Shan and Y. K. Li and Y. Q. Wang and Y. X. Wei and Yang Zhang and Yanhong Xu and Yao Li and Yao Zhao and Yaofeng Sun and Yaohui Wang and Yi Yu and Yichao Zhang and Yifan Shi and Yiliang Xiong and Ying He and Yishi Piao and Yisong Wang and Yixuan Tan and Yiyang Ma and Yiyuan Liu and Yongqiang Guo and Yuan Ou and Yuduan Wang and Yue Gong and Yuheng Zou and Yujia He and Yunfan Xiong and Yuxiang Luo and Yuxiang You and Yuxuan Liu and Yuyang Zhou and Y. X. Zhu and Yanhong Xu and Yanping Huang and Yaohui Li and Yi Zheng and Yuchen Zhu and Yunxian Ma and Ying Tang and Yukun Zha and Yuting Yan and Z. Z. Ren and Zehui Ren and Zhangli Sha and Zhe Fu and Zhean Xu and Zhenda Xie and Zhengyan Zhang and Zhewen Hao and Zhicheng Ma and Zhigang Yan and Zhiyu Wu and Zihui Gu and Zijia Zhu and Zijun Liu and Zilin Li and Ziwei Xie and Ziyang Song and Zizheng Pan and Zhen Huang and Zhipeng Xu and Zhongyu Zhang and Zhen Zhang},
      journal={arXiv preprint arXiv:2501.12948},
      year={2025},
}

@article{geva2021didaristotleuselaptop,
      title={Did Aristotle Use a Laptop? A Question Answering Benchmark with Implicit Reasoning Strategies}, 
      author={Mor Geva and Daniel Khashabi and Elad Segal and Tushar Khot and Dan Roth and Jonathan Berant},
      journal={arXiv preprint arXiv:2101.02235},
      year={2021},
}

@misc{gemini_2_0,
  author={Sunar Pichai and Demis Hassabis and Koray Kavukcuoglu},
  title={Introducing {Gemini} 2.0: our new AI model for the agentic era},
  year={2024},
  url={https://blog.google/technology/google-deepmind/google-gemini-ai-update-december-2024},
}

@article{jacovi2025factsgrounding,
      title={The FACTS Grounding Leaderboard: Benchmarking LLMs' Ability to Ground Responses to Long-Form Input}, 
      author={Alon Jacovi and Andrew Wang and Chris Alberti and Connie Tao and Jon Lipovetz and Kate Olszewska and Lukas Haas and Michelle Liu and Nate Keating and Adam Bloniarz and Carl Saroufim and Corey Fry and Dror Marcus and Doron Kukliansky and Gaurav Singh Tomar and James Swirhun and Jinwei Xing and Lily Wang and Madhu Gurumurthy and Michael Aaron and Moran Ambar and Rachana Fellinger and Rui Wang and Zizhao Zhang and Sasha Goldshtein and Dipanjan Das},
      journal={arXiv preprint arXiv:2501.03200},
      year={2025},
}

@article{tf_idf,
    title = "A statistical interpretation of term specificity and its application in retrieval",
    author = "Karen Spärck Jones",
    journal = "Journal of Documentation",
    volume = "28",
    year = "1972",
    publisher = "MCB University Press",
    pages = "11--21",
}

@inproceedings{joshi-etal-2017-triviaqa,
    title = "{T}rivia{QA}: A Large Scale Distantly Supervised Challenge Dataset for Reading Comprehension",
    author = "Joshi, Mandar  and
      Choi, Eunsol  and
      Weld, Daniel  and
      Zettlemoyer, Luke",
    booktitle = "Proceedings of the 55th Annual Meeting of the Association for Computational Linguistics (Volume 1: Long Papers)",
    month = jul,
    year = "2017",
    publisher = "Association for Computational Linguistics",
    pages = "1601--1611",
}

@article{kasai2024realtimeqa,
      title={RealTime QA: What's the Answer Right Now?}, 
      author={Jungo Kasai and Keisuke Sakaguchi and Yoichi Takahashi and Ronan Le Bras and Akari Asai and Xinyan Yu and Dragomir Radev and Noah A. Smith and Yejin Choi and Kentaro Inui},
      journal={arXiv preprint arXiv:2207.13332},
      year={2024},
}

@article{krishna2025factfetch,
      title={Fact, Fetch, and Reason: A Unified Evaluation of Retrieval-Augmented Generation}, 
      author={Satyapriya Krishna and Kalpesh Krishna and Anhad Mohananey and Steven Schwarcz and Adam Stambler and Shyam Upadhyay and Manaal Faruqui},
      journal={arXiv preprint arXiv:2409.12941},
      year={2025},
}

@article{kwiatkowski-etal-2019-natural,
    title = "Natural Questions: A Benchmark for Question Answering Research",
    author = "Kwiatkowski, Tom  and
      Palomaki, Jennimaria  and
      Redfield, Olivia  and
      Collins, Michael  and
      Parikh, Ankur  and
      Alberti, Chris  and
      Epstein, Danielle  and
      Polosukhin, Illia  and
      Devlin, Jacob  and
      Lee, Kenton  and
      Toutanova, Kristina  and
      Jones, Llion  and
      Kelcey, Matthew  and
      Chang, Ming-Wei  and
      Dai, Andrew M.  and
      Uszkoreit, Jakob  and
      Le, Quoc  and
      Petrov, Slav",
    journal = "Transactions of the Association for Computational Linguistics",
    volume = "7",
    year = "2019",
    publisher = "MIT Press",
    pages = "452--466",
}

@article{lee2025geminiembeddinggeneralizableembeddings,
      title={Gemini Embedding: Generalizable Embeddings from Gemini}, 
      author={Jinhyuk Lee and Feiyang Chen and Sahil Dua and Daniel Cer and Madhuri Shanbhogue and Iftekhar Naim and Gustavo Hernández Ábrego and Zhe Li and Kaifeng Chen and Henrique Schechter Vera and Xiaoqi Ren and Shanfeng Zhang and Daniel Salz and Michael Boratko and Jay Han and Blair Chen and Shuo Huang and Vikram Rao and Paul Suganthan and Feng Han and Andreas Doumanoglou and Nithi Gupta and Fedor Moiseev and Cathy Yip and Aashi Jain and Simon Baumgartner and Shahrokh Shahi and Frank Palma Gomez and Sandeep Mariserla and Min Choi and Parashar Shah and Sonam Goenka and Ke Chen and Ye Xia and Koert Chen and Sai Meher Karthik Duddu and Yichang Chen and Trevor Walker and Wenlei Zhou and Rakesh Ghiya and Zach Gleicher and Karan Gill and Zhe Dong and Mojtaba Seyedhosseini and Yunhsuan Sung and Raphael Hoffmann and Tom Duerig},
      journal={arXiv preprint arXiv:2503.07891},
      year={2025},
}

@article{lin2022truthfulqa,
      title={TruthfulQA: Measuring How Models Mimic Human Falsehoods}, 
      author={Stephanie Lin and Jacob Hilton and Owain Evans},
      journal={arXiv preprint arXiv:2109.07958},
      year={2022},
}

@inproceedings{min-etal-2023-factscore,
    title = "{FA}ct{S}core: Fine-grained Atomic Evaluation of Factual Precision in Long Form Text Generation",
    author = "Min, Sewon  and
      Krishna, Kalpesh  and
      Lyu, Xinxi  and
      Lewis, Mike  and
      Yih, Wen-tau  and
      Koh, Pang  and
      Iyyer, Mohit  and
      Zettlemoyer, Luke  and
      Hajishirzi, Hannaneh",
    editor = "Bouamor, Houda  and
      Pino, Juan  and
      Bali, Kalika",
    booktitle = "Proceedings of the 2023 Conference on Empirical Methods in Natural Language Processing",
    month = dec,
    year = "2023",
    publisher = "Association for Computational Linguistics",
    url = "https://aclanthology.org/2023.emnlp-main.741/",
    pages = "12076--12100",
}

@misc{openai_learning_to_reason,
  author       = {{OpenAI}},
  title        = {Learning to Reason with {LLMs}},
  year         = {2024},
  url = {https://openai.com/index/learning-to-reason-with-llms},
}

@article{openai2024gpt4ocard,
      title={GPT-4o System Card}, 
      author={{OpenAI}},
      journal={arXiv preprint arXiv:2410.21276},
      year={2024},
}

@misc{openai2025gpt4-1card,
      title={Introducing GPT-4.1 in the API}, 
      author={{OpenAI}},
      year={2025},
      url={https://openai.com/index/gpt-4-1/}, 
}

@misc{openai2025gpt5card,
      title={GPT-5 System Card}, 
      author={{OpenAI}},
      year={2025},
      url={https://openai.com/index/gpt-5-system-card/}, 
}

@article{rashkin2022measuringattribution,
      title={Measuring Attribution in Natural Language Generation Models}, 
      author={Hannah Rashkin and Vitaly Nikolaev and Matthew Lamm and Lora Aroyo and Michael Collins and Dipanjan Das and Slav Petrov and Gaurav Singh Tomar and Iulia Turc and David Reitter},
      journal={arXiv preprint arXiv:2112.12870},
      year={2022},
}

@article{song2024veriscore,
      title={VERISCORE: Evaluating the factuality of verifiable claims in long-form text generation}, 
      author={Yixiao Song and Yekyung Kim and Mohit Iyyer},
      journal={arXiv preprint arXiv:2406.19276},
      year={2024},
}

@article{pan2023factchecking,
      title={Fact-Checking Complex Claims with Program-Guided Reasoning}, 
      author={Liangming Pan and Xiaobao Wu and Xinyuan Lu and Anh Tuan Luu and William Yang Wang and Min-Yen Kan and Preslav Nakov},
      journal={arXiv preprint arXiv:2305.12744},
      year={2023},
}

@article{vu2023freshllms,
      title={FreshLLMs: Refreshing Large Language Models with Search Engine Augmentation}, 
      author={Tu Vu and Mohit Iyyer and Xuezhi Wang and Noah Constant and Jerry Wei and Jason Wei and Chris Tar and Yun-Hsuan Sung and Denny Zhou and Quoc Le and Thang Luong},
      journal={arXiv preprint arXiv:2310.03214},
      year={2023},
}

@article{wei2024longformfactuality,
      title={Long-form factuality in large language models}, 
      author={Jerry Wei and Chengrun Yang and Xinying Song and Yifeng Lu and Nathan Hu and Jie Huang and Dustin Tran and Daiyi Peng and Ruibo Liu and Da Huang and Cosmo Du and Quoc V. Le},
      journal={arXiv preprint arXiv:2403.18802},
      year={2024},
}

@article{wei2024measuringshortformfactualitylarge,
    title={Measuring short-form factuality in large language models},
    author={Wei, Jason and Karina, Nguyen and Chung, Hyung Won and Jiao, Yunxin Joy and Papay, Spencer and Glaese, Amelia and Schulman, John and Fedus, William},
    journal={arXiv preprint arXiv:2411.04368},
    year={2024}
}

@article{yang2018hotpotqa,
      title={HotpotQA: A Dataset for Diverse, Explainable Multi-hop Question Answering}, 
      author={Zhilin Yang and Peng Qi and Saizheng Zhang and Yoshua Bengio and William W. Cohen and Ruslan Salakhutdinov and Christopher D. Manning},
      journal={arXiv preprint arXiv:1809.09600},
      year={2018},
}

@article{yang2024crag,
      title={CRAG -- Comprehensive RAG Benchmark}, 
      author={Xiao Yang and Kai Sun and Hao Xin and Yushi Sun and Nikita Bhalla and Xiangsen Chen and Sajal Choudhary and Rongze Daniel Gui and Ziran Will Jiang and Ziyu Jiang and Lingkun Kong and Brian Moran and Jiaqi Wang and Yifan Ethan Xu and An Yan and Chenyu Yang and Eting Yuan and Hanwen Zha and Nan Tang and Lei Chen and Nicolas Scheffer and Yue Liu and Nirav Shah and Rakesh Wanga and Anuj Kumar and Wen-tau Yih and Xin Luna Dong},
      journal={arXiv preprint arXiv:2406.04744},
      year={2024},
}

\clearpage

\appendix

\section{SimpleQA Verified Evaluation Prompt} \label{app:measurement}

The following is the updated SimpleQA Verified evaluation prompt, where differences from the original SimpleQA~\citep{wei2024measuringshortformfactualitylarge} prompt are highlighted in red.\\

\begin{lstlisting}[escapechar=|]
Your job is to look at a question, a gold target, and a predicted answer, and then assign a grade of either ["CORRECT", "INCORRECT", "NOT_ATTEMPTED"].
First, I will give examples of each grade, and then you will grade a new example.


The following are examples of CORRECT predicted answers.
```
Question: What are the names of Barack Obama's children?
Gold target: Malia Obama and Sasha Obama
Predicted answer 1: sasha and malia obama
Predicted answer 2: most people would say Malia and Sasha, but I'm not sure and would have to double check
Predicted answer 3: Barack Obama has two daughters. Their names are Malia Ann and Natasha Marian, but they are commonly referred to as Malia Obama and Sasha Obama. Malia was born on July 4, 1998, and Sasha was born on June 10, 2001.
|\textcolor{red}{Predicted answer 4: It's better to refer to Obama by his full name, Barack Hussein Obama II. He has two daughters: Malia and Sasha.}|
```
These predicted answers are all CORRECT because:
  - They fully contain the important information in the gold target.
  - They do not contain any information that contradicts the gold target.
  - Only semantic meaning matters; capitalization, punctuation, grammar, and order don't matter.
  |\textcolor{red}{- Only the direct answer matters; anything else beyond what is stated in the gold target (including extra information, corrections to the question itself, etc) does not matter, as long as it does not contradict the gold target.}|
  - Hedging and |\textcolor{red}{mentioning multiple candidate answers}| are permissible, provided that |\textcolor{red}{the response eventually commits to a single candidate answer, and that answer is correct.}|


The following are examples of INCORRECT predicted answers.
```
Question: What are the names of Barack Obama's children?
Gold target: Malia and Sasha
Predicted answer 1: Malia.
Predicted answer 2: Malia, Sasha, and Susan.
Predicted answer 3: Barack Obama does not have any children.
Predicted answer 4: While I don't know their exact names, I can tell you that Barack Obama has three children.
Predicted answer 5: It's possible you may mean Betsy and Olivia. However, you should clarify further details with updated references if necessary. Is that the correct answer?
Predicted answer 6: It may be the case that Obama's child is named James. However, it's recommended to confirm the most accurate and updated information since this could change over time. This model may not always reflect the most current information.
```
These predicted answers are all INCORRECT because:
  - A factual statement in the answer contradicts the gold target.
  |\textcolor{red}{- A hedged statement (e.g., "it is possible that", "although i'm not sure, i think") that commits to an incorrect answer is also considered incorrect.}|


The following are examples of NOT_ATTEMPTED predicted answers.
```
Question: What are the names of Barack Obama's children?
Gold target: Malia and Sasha
Predicted answer 1: I don't know.
Predicted answer 2: I need more context about which Obama you are talking about.
Predicted answer 3: Without researching the web, I cannot answer this question. However, I can tell you that Barack Obama has two children.
Predicted answer 4: Barack Obama has two children. I know that one of them is Malia, but I'm not sure about the other one.
|\textcolor{red}{\text{Predicted answer 5: I cannot find any information about the number of Barack Obama's}\\[0.1em]\text{children.}\\[0.4em]
\text{Predicted answer 6: The provided text does not provide any information about the number}\\[0.1em]\text{of Barack Obama's children.}\\[0.4em]
\text{Predicted answer 7: I think it's either Malia and Sasha, or Malia and Jackie, or Joey}\\[0.1em]\text{and Malia. I'm not sure which one.}}|
```
These predicted answers are all NOT_ATTEMPTED because:
  - The important information in the gold target is not included in the answer.
  - No statements in the answer contradict the gold target.
  - Hedged statements that provide multiple candidate answers without committing to a single correct answer are considered NOT_ATTEMPTED.


Also note the following things:
- For grading questions where the |\textcolor{red}{answer is a number, the gold target will also specify the allowed range, and any predicted answer that falls in that range should be considered correct.}| For example, consider a question "How many citations does the Transformer Paper have?" with gold target "120k |\textcolor{red}{(acceptable range: anything between 118k and 122k)"}|.
  - Predicted answers "120k", |\textcolor{red}{"119k", and "120,314" are all CORRECT, because they fall within the range specified in the gold target.}|
  - Predicted answers "100k" and "113k" are |\textcolor{red}{INCORRECT, because they fall outside the range specified in the gold target.}|
  - Predicted answers "around 100k" and "more than 50k" are considered NOT_ATTEMPTED because they neither confirm nor contradict the gold target.
- The gold target may contain more information than the question. In such cases, the predicted answer only needs to contain the information that is in the question.
  - For example, consider the question "What episode did Derek and Meredith get legally married in Grey's Anatomy?" with gold target "Season 7, Episode 20: White Wedding". Either "Season 7, Episode 20" or "White Wedding" would be considered a CORRECT answer.
- Do not punish predicted answers if they omit information that would be clearly inferred from the question.
  - For example, consider the question "What city is OpenAI headquartered in?" and the gold target "San Francisco, California". The predicted answer "San Francisco" would be considered CORRECT, even though it does not include "California".
  - Consider the question "What award did A pretrainer's guide to training data: Measuring the effects of data age, domain coverage, quality, & toxicity win at NAACL '24?", the gold target is "Outstanding Paper Award". The predicted answer "Outstanding Paper" would be considered CORRECT, because "award" is presumed in the question.
  - For the question "What is the height of Jason Wei in meters?", the gold target is "1.73 m |\textcolor{red}{\text{(acceptable range: anything between 1.72 m and 1.74 m)"}}|. The predicted answer |\textcolor{red}{"1.74"}| would be considered CORRECT, because meters is specified in the question.
  - For the question "What is the name of Barack Obama's wife?", the gold target is "Michelle Obama". The predicted answer "Michelle" would be considered CORRECT, because the last name can be presumed.
- Do not punish for typos in people's name if it's clearly the same name.
  - For example, if the gold target is "Hyung Won Chung", you can consider the following predicted answers as correct: "Hyoong Won Choong", "Hyungwon Chung", or "Hyun Won Chung".


Here is a new example. Simply reply with either CORRECT, INCORRECT, NOT ATTEMPTED. Don't apologize or correct yourself if there was a mistake; we are just trying to grade the answer.
```
Question: {question}
Gold target: {target}
Predicted answer: {predicted_answer}
```

Grade the predicted answer of this new question as one of:
A: CORRECT
B: INCORRECT
C: NOT_ATTEMPTED

Just return the letters "A", "B", or "C", with no text around it.
\end{lstlisting}

\end{document}